\documentclass[runningheads]{llncs}

\usepackage[mobile]{eccv}

\usepackage{eccvabbrv}

\usepackage{graphicx}
\usepackage{booktabs}

\usepackage[accsupp]{axessibility}  

\usepackage{url}
\usepackage{orcidlink}

\usepackage{comment}
\usepackage{ifthen} 
\usepackage{overpic}
\usepackage{booktabs}
\usepackage{multirow}
\usepackage{rotating}
\usepackage{wrapfig}
\usepackage{amsmath}
\usepackage{amssymb}
\usepackage{bbm}
\usepackage{dsfont}
\usepackage{wrapfig}
\usepackage{verbatim} 
\usepackage{nicefrac} 
\usepackage{caption}
\usepackage{pifont}
\usepackage{colortbl} 
\usepackage{color}
\usepackage{xcolor}
\usepackage{balance}
\usepackage{marvosym}

\usepackage{float}

\usepackage{textcomp}

\newcommand{\mathmat}[1]{\mathbf{#1}}
\newcommand{\mathset}[1]{\mathcal{#1}}

\newcommand{\secref}[1]{Section \ref{#1}}
\newcommand{\tabref}[1]{Table \ref{#1}}
\newcommand{\figref}[1]{Figure \ref{#1}}

\makeatletter
\def\thanks#1{\protected@xdef\@thanks{\@thanks
        \protect\footnotetext{#1}}}
\makeatother

\begin{document}



\title{Graphic Design with Large Multimodal Model}


\titlerunning{HLG}

\author{
Yutao Cheng\textcolor{magenta}{*} \inst{1}\orcidlink{0009-0002-3618-1344} \and
Zhao Zhang\textcolor{magenta}{*} \inst{1}\orcidlink{0000-0002-1521-8163} \and
Maoke Yang\textcolor{magenta}{*} \inst{1}\orcidlink{0000-0002-5400-4117} \and \\
Hui Nie \inst{1,2} \and
Chunyuan Li \inst{1} \and
Xinglong Wu \inst{1} \and
Jie Shao\textsuperscript{\Letter} \inst{1} 
}

\authorrunning{Graphist}

\institute{
{\small$^1$ByteDance \quad $^2$UCAS} \\
\email{\{yutao.135,yangmaoke,shaojie.mail\}@bytedance.com} \\
\email{zzhang@mail.nankai.edu.cn}
\thanks{
\textcolor{magenta}{*} Equal Contributions
}
}

\maketitle

\begin{abstract}
In the field of graphic design, automating the integration of design elements into a cohesive multi-layered artwork not only boosts productivity but also paves the way for the democratization of graphic design.
One existing practice is Graphic Layout Generation (GLG), which aims to layout sequential design elements. 
It has been constrained by the necessity for a predefined correct sequence of layers, thus limiting creative potential and increasing user workload. 
In this paper, we present Hierarchical Layout Generation (HLG) as a more flexible and pragmatic setup, which creates graphic composition from \textit{\textbf{unordered}} sets of design elements.
To tackle the HLG task, we introduce Graphist, the first layout generation model based on  large multimodal models. 
Graphist efficiently reframes the HLG as a sequence generation problem, utilizing RGB-A images as input, outputs a JSON draft protocol, indicating the coordinates, size, and order of each element. 
We develop multiple evaluation metrics for HLG.
Graphist outperforms prior arts and establishes a strong baseline for this field.
Project homepage: \url{https://github.com/graphic-design-ai/graphist}
\end{abstract}
\begin{figure}[ht!]
\centering
\begin{overpic}[width=0.99\textwidth]{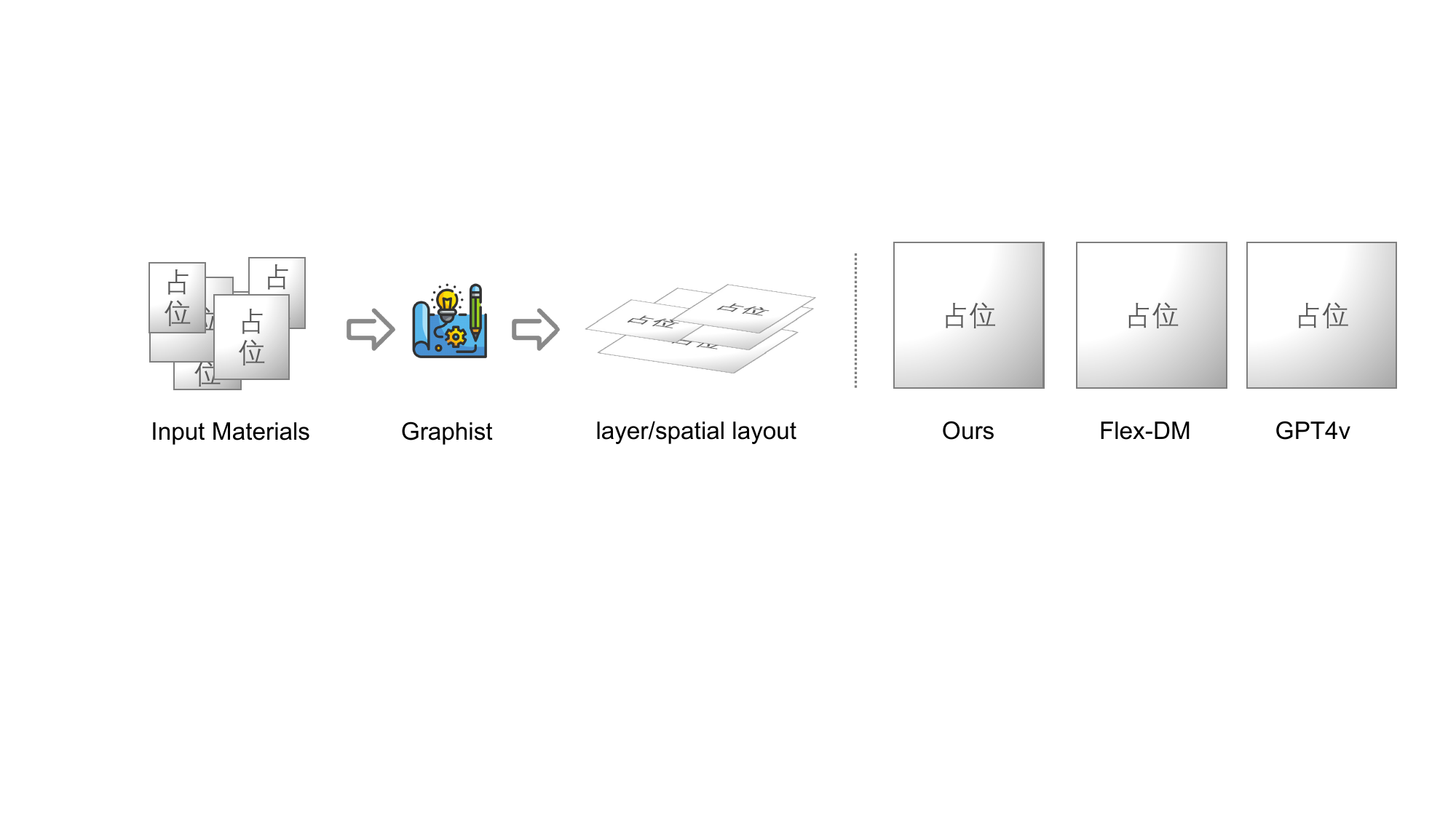}
\end{overpic}
\caption{
\textbf{Schematic diagram of hierarchical layout generation.}
(Left) A comparison between the traditional \textit{GLG} task and the newly proposed \textit{HLG} task, with the major difference in that HLG relaxes the constraint of GLG, so that unordered multimodal input elements can be processed. (Right) Errors in either layer sequencing or spatial positioning can significantly impact the overall quality of the design.
}
\label{fig:teaser}
\end{figure}

\section{Introduction}
\label{sec:intro}
Graphic design~\cite{jobling1996graphic_design} fundamentally serves as a form of visual communication. It involves the creation and combination of symbols, images, and text to express certain ideas or messages. This field requires significant expertise and time investment to produce aesthetically pleasing graphic compositions.
Recently, there is a significant paradigm shift in leveraging AI for automating the layout of given design elements into cohesive graphic compositions heralds~\cite{zhong2019publaynet,yin2013social_media,ganin2021computer,fu2022doc2ppt}. This could potentially reduce the workload of professional designers and provide an avenue for beginners to create their design pieces, making graphic design more democratize and effective.

A preliminary attempt to automate this process is observed in the task known as \textbf{G}raphic \textbf{L}ayout \textbf{G}eneration (GLG)~\cite{yamaguchi2021crello,jiang2022coarse,feng2023layoutgpt,lin2023layoutprompter,inoue2023layout,zhang2023layoutdiffusion,lin2023parse,jiang2023layoutformer++,weng2023learn,cheng2023play,hsu2023posterlayout,li2023planning}. 
GLG attempts to intelligently arrange provided elements into attractive compositions under the assumption of a \textit{predefined order of layers}. 
However, establishing the appropriate ordering of these layers is a design cornerstone that, if mismanaged, can fracture the visual hierarchy, leading to disarray in the intended message delivery. 
Requiring users to prescribe an accurate layer sequence prior to layout not only burdens them with foresight and planning but also stifles layout algorithms, restricting their capacity to transcend such confines in the pursuit of innovative and aesthetically superior outcomes.

In pursuit of more practical applications, this paper introduces a new task, \textbf{H}ierarchical \textbf{L}ayout \textbf{G}eneration (HLG). 
HLG to craft a visually appealing graphic composition from a collection of unordered elements by meticulously considering both their spatial arrangement and the sequencing of layers. 
The adjective \textit{\textbf{hierarchical}} emphasizes the significance of element ordering, setting HLG apart from conventional layout generation practices. 
The distinction between GLG and HLG is illustrated in the left panel of \figref{fig:teaser}, while the right panel of the same figure captures the impact of layer sequencing and spatial positioning on the visual aesthetics of graphic design.

\begin{figure}[t]
\begin{overpic}[width=0.99\textwidth]{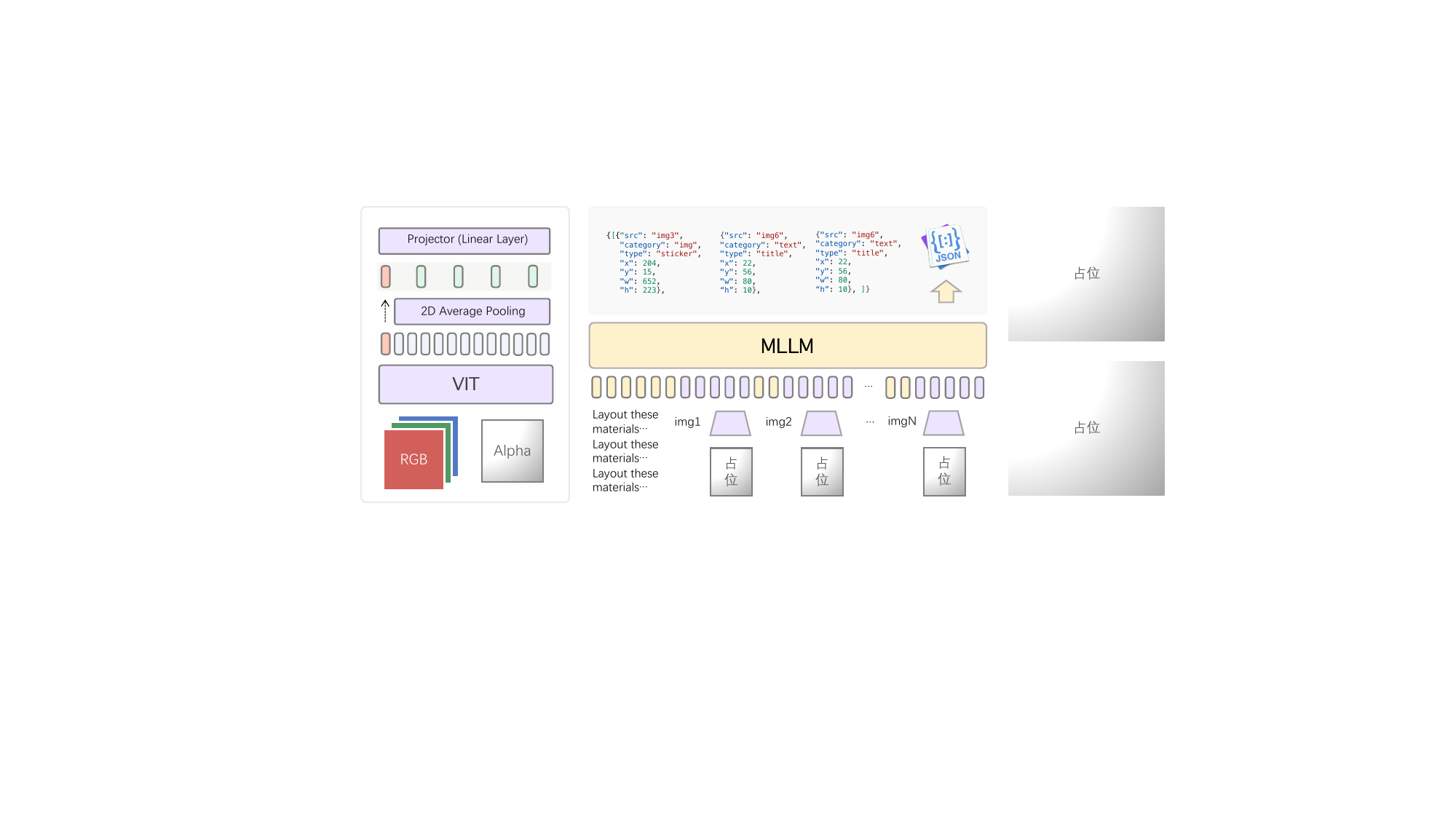}
\end{overpic}
\caption{\textbf{Graphist Pipeline.} Graphist comprises three components: RGB-A Encoder, Visual Shrinker, and a LLM. It accepts a variety of design elements and generates a graphic composition in JSON format end-to-end.}
\label{fig:pipeline}
\end{figure}

For the HLG task, we introduce Graphist, the first layout generation model built upon Large Multimodal Model (LMM)~\cite{li2023blip2, dai2023instructblip, liu2023llava, zhu2023minigpt, chen2023shikra,tai2023link}. 
The layout generation task is challenging that it digests diverse input elements such as RGB-A materials, RGB images, and texts, and the desired outcomes must precisely reflect the intricate relationships among these multimodal input elements. 
LMMs are well-suited for this task, as they can unify different modalities, like images, text, coordinates~\cite{peng2023kosmos2,chen2023shikra} into tokens.
This allows for flexible configuration of various tasks, such as HLG, GLG, and more variants.
The knowledge and strong reasoning ability stored in pre-trained model parameters are also beneficial for layout tasks.
Furthermore, LMMs demonstrate significant potential for scaling~\cite{kaplan2020scaling,openai2023gpt4,team2023gemini}, enabling the pursuit of enhanced performance through the use of larger models and more extensive datasets.
For these reasons, LMMs were a natural choice for our foundational architecture in developing Graphist.
In our specific approach for HLG,
we train Graphist with graphic composition data, in which, as shown in \figref{fig:pipeline},
each design element is represented as an RGB-A image as input,
the model then generates a JSON draft protocol, which specifies the coordinates, size, order and other attributes of each design elements in end-to-end manner.

To evaluate the HLG task, 
we introduce two novel metrics: Inverse Order Pair Ratio (IOPR) and GPT-4V Eval. 
The former assesses the accuracy of the layer order in the graphic composition, 
while the GPT-4V Eval, leveraging the capabilities of GPT-4V~\cite{openai2023gpt4v}, quantifies the overall aesthetic quality.
Additionally, we have incorporated a human rating score to align with the subjective perceptions of real individuals. Graphist has emerged as the SoTA solution, excelling across all these metrics. Furthermore, Graphist consistently delivers impressive performance on conventional GLG task.

We summarize our contributions as follows:
\begin{itemize}
    \item We introduce the Hierarchical Layout Generation (HLG) task, which creates graphic compositions from unordered design elements. HLG overcomes the constraints of the Graphic Layout Generation that requires pre-determined layer ordering. The new setup allows more flexible and practical AI-assisted graphic designs.
    \item We present Graphist, the first  LMM-parameterized layout generation model that can be trained end-to-end. 
    Graphist accepts a variety of vision-and-language multimodal design elements and generates a graphic compositions in JSON code format, which can be further automatically rendered into the final design.
    \item We develop evaluation metrics for HLG, including IOPR and GPT-4V Eval. Graphist shows superior performance in these metrics, setting a robust benchmark for the field.
\end{itemize}

\section{Related Work}
\label{sec:related_work}
\begin{figure}[t]
\centering
\begin{overpic}[width=1.0\textwidth]{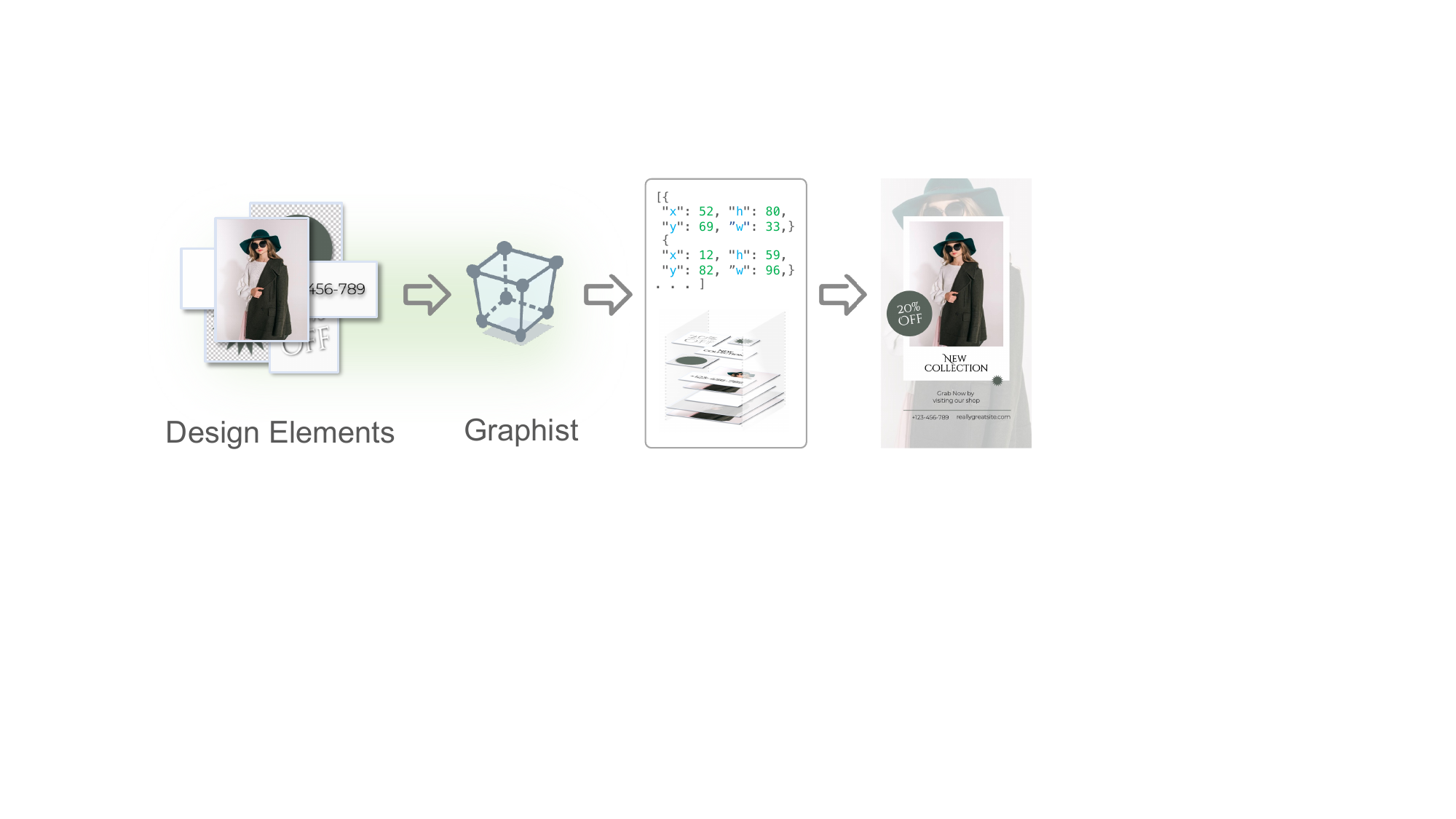}
\end{overpic}
\caption{
\textbf{A user-generated case via graphist web demo}.
 The top-left figure represents the input design elements to Graphist. Below it, we present the corresponding output JSON code generated by Graphist. The final two images in the top row illustrate the visualized results: first is the layout visualization, and the second is the graphic composition by putting these elements according to the JSON protocol.
 Additional examples are available in \figref{fig:user}.
}
\label{fig:one_case}
\end{figure}

\subsection{Graphic Layout Generation}
Graphic design is a form of visual art that combines multimodal elements (\eg, images, texts, and symbols) to create aesthetically pleasing compositions which can effectively convey information to the audience. 
As one of the core technologies in automated graphic design, layout generation methods has been widely used in various scenarios, such as document~\cite{zhong2019publaynet, jiang2023layoutformer++}, UI~\cite{deka2017rico, beltramelli2018pix2code,lu2023uilayout, wei2023boosting}, social media snippets~\cite{yin2013social_media}, banners~\cite{lee2020ndn}, poster and advertisement~\cite{hsu2023posterlayout, li2023planning, lin2023autoposter}, icon/logo~\cite{carlier2020deepsvg,feng2022autoicon,wu2023iconshop,mateja2023animatesvg}, CAD Sketches~\cite{para2021sketchgen,ganin2021computer}, slides~\cite{shibata2005automatic,hu2014ppsgen,fu2022doc2ppt}.

As people deepen their understanding of layout generation tasks, the modeling of layout generation problems is becoming increasingly complex. 
Early methods for layout generation typically relied on aesthetic rules~\cite{bauerly2006computational_aesthetics, yang2016automatic} or constraints~\cite{hurst2009review_layout, jahanian2013recommendation_magazine}.
However, the diversity and aesthetics of generated compositions are primarily restricted by the formalized human prior knowledge, and are hard to scale up for production. 
Later on, some data-driven methods~\cite{li2019layoutgan, lee2020ndn, arroyo2021variational} emerged, 
utilizing learning-based generation models for layout generation. 
These approaches are dedicated to fitting the data distribution of the topological relationship between elements, and focus on the alignment of elements, 
but neglecting the content of the elements. 
Recent methods~\cite{zhou2022composition, yu2022layoutdetr, chai2023two, gao2023textpainter, li2023relation, lin2023autoposter} have recognized the importance of element content, 
and have explored ways to layout text or decorations on a given base graphic without obstructing important content. 
Drawing from this observation, attempts~\cite{inoue2023FlexDM, zheng2019content} have also been made to incorporate multimodal inputs in order to better understand the content of each layer. 
These methods often assume that the layering order of the elements has been predefined~\cite{yamaguchi2021crello, inoue2023FlexDM}, which is difficult to achieve in the practical application. 
Therefore, we argue that the layer ordering needs to be fully considered in the layout generation problem.

The advancement of layout generation techniques is closely intertwined with the progress of fundamental algorithms. During the early exploration~\cite{bauerly2006computational_aesthetics, yang2016automatic, hurst2009review_layout, jahanian2013recommendation_magazine}, researchers attempted to define beauty using formulas and constraints. 
Subsequently, generative methods based on GANs~\cite{li2019layoutgan, li2020attribute, hsu2023densitylayout}, VAEs~\cite{lee2020ndn, arroyo2021variational}, and diffusions~\cite{zhang2023layoutdiffusion, li2023relation, wei2023boosting} have been widely attempted. 
Besides, LayoutDETR~\cite{yu2022layoutdetr} combining DETR~\cite{carion2020detr} and generative models has been attempted to deal with the problem of element arrangement on a given image. 
In the era of Transformer~\cite{vaswani2017attention}, 
several methods~\cite{inoue2023FlexDM, jiang2023layoutformer++, lin2023parse} have adopted the powerful learning capabilities of Transformers to address layout and related control problems~\cite{jiang2023layoutformer++, lin2023parse}. 
FlexDM~\cite{inoue2023FlexDM} models graphic design using a BERT-like~\cite{devlin2018bert} approach. 
The rise of large language models~\cite{brown2020gpt3, touvron2023llama, touvron2023llama2} has also captured the attention of researchers in the field of graphic design, with LayoutPrompter~\cite{lin2023layoutprompter} and LayoutNUWA~\cite{tang2023layoutnuwa} introducing language models into layout generation in a zero-shot manner. 
COLE~\cite{jia2023cole} fuses large language models and diffusion models to build a hierarchical graphic design framework. 
Our approach is also inspired by large language models~\cite{touvron2023llama2, zhang2024tinyllama} and multimodal models~\cite{liu2023llava, liu2023llava1_5, zhu2023minigpt, chen2023shikra}. 
However, COLE~\cite{jia2023cole} focuses on utilizing the generating power of diffusion models to directly produce multi-layer graphic designs based on user intent, while our approach prioritizes the use of existing design elements for layout generation.

\subsection{Large Multimodal Models}
As a bridge between language and vision,  LMM~\cite{li2023multimodal,yin2023survey,zhang2024mm} has received widespread attention recently. 
Related topic including autonomous driving \cite{ding2024holistic,cui2024survey}, video understanding~\cite{tang2023video, zhang2023videollama, li2023videochat}, open-vocabulary vision understanding~\cite{yuan2023osprey,wu2024open_vocabulary, omgseg}, multimodal agent~\cite{yang2023appagent,wang2024mobileagent}, image generation~\cite{betker2023dalle3,sun2023generative,zhan2024anygpt}, embodied intelligence~\cite{driess2023palme,mu2024embodiedgpt,brohan2023rt} and so on. To the best of our knowledge, Graphist is the first work to address the task of layout generation using LMM in an end-to-end manner. While COLE~\cite{jia2023cole} likewise employs multimodal models in their pipeline, their main emphasis lies in generating text content and styles on predetermined base image.

Common architectures of LMM typically encompass a pre-trained visual encoder~\cite{dosovitskiy2020vit, radford2021learning} for extracting visual features, a pre-trained LLM\cite{touvron2023llama, touvron2023llama2, 2023internlm} for interpreting user commands and generating responses, as well as a vision-language cross-modal connector\cite{alayrac2022flamingo, liu2023llava, li2023blip2, gao2023llama_adapter, zhou2023infmllm} for aligning the visual encoder outputs to the language model.
Given that the task of hierarchical layout generation involves organizing design elements on a blank canvas, the compatibility of coordinates is of paramount importance.
Pix2Seq~\cite{chen2021pix2seq} was the first to explore the use of a discretization and serialization approach in object detection problems, in order to convert coordinates into token sequences that can ultimately be used in sequence generation mode. 
Some work, like OFA~\cite{wang2022ofa}, VisionLLM~\cite{wang2023visionllm}, Kosmos-2~\cite{kosmos}, build upon this form by introducing special coordinate tokens in their vocabulary.
Alternatively, 
PerceptionGPT~\cite{pi2023perceptiongpt} implements an additional vision encoder and decoder specifically for processing and predicting coordinates.
Most notable among these approaches is Shikra~\cite{chen2023shikra}, which represents spatial positions using numerical values in natural language.
It has been proven succinct and effective, as echoed by related research~\cite{bai2023qwenvl,you2023ferret,wang2023cogvlm}. Inspired by Shikra, our work adopts the numerical representation for coordinates in the natural language sequence.


\section{Task Formulation}
\subsection{Graphic Layout Generation}
For a unified representation, we assume that a text box can be represented by an RGB-A image, which can usually be achieved through rendering or similar methods. Under this assumption, GLG is defined as finding an optimal arrangement for a collection of RGBA design assets $\mathset{M} = \{M_{i}\in\mathbb{R}^{h_{i}\times w_{i} \times 4}\}_{i=1}^{n}$. This endeavor seeks to spatially organize the set in a coherent graphic composition $\mathset{S}=\{s_{i}\}_{i=1}^{n}=f(\mathset{M})$, where $f(\cdot)$ embodies our Graphist layout methodology. Specifically, the output constitutes a set of transformations wherein for each asset $M_i$, a quadruple $(x_i, y_i, w_i, h_i)$ expresses its upper-left corner placement along with the respective width and height adjustments within the composition.

\subsection{Hierarchical Layout Generation}
Given a set of RGB-A design elements $\mathset{M} = \{M_{i}\in\mathbb{R}^{h_{i}\times w_{i} \times 4}\}_{i=1}^{n}$, HLG seeks to arrange them into a well-constructed graphic composition $\mathset{S}=\{s_{i}\}_{i=1}^{n}=f(\mathset{M})$. In this context, $f(\cdot)$ denotes the layout action implemented by our method, which we refer to as Graphist. Each element $s_i$ in the layout prediction encompasses five numerical values $(x_i, y_i, w_i, h_i, l_i)$, associated with $M_{i}$ top/left coordinates, width, height and hierarchy respectively.
As shown in \figref{fig:teaser}, compare to conventional GLG task~\cite{yamaguchi2021crello},
HLG framework emphasizes the significance of element stratification by imposing a hierarchy $l_i$ that dictates not only the placement but also the $z$-order of each element, thus ensuring a compositionally harmonious layering.

\section{Proposed Method}
\label{sec:method}
\subsection{Graphist Architecture}
Graphist is constructed using a LMM, which receives the input of multimodal design elements and predicts a JSON fragment formatted like \figref{fig:pipeline}.
Graphist comprises three components: 
$(i)$
{\it  RGBA-Encoder}.
We utilize a ViT-L/14 with $224 \times 224$ four-channel input as RGBA-Encoder, initialized with visual tower parameters of CLIP~\cite{radford2021learning}, excluding the alpha channel.
$(ii)$ 
{\it LLM}.
Our LLM foundation incorporates Qwen1.5-0.5B/7B~\cite{qwen}.
$(iii)$
{\it Visual Shrinker}.
To manage the extensive elements processing requirements of Graphist, the Visual Shrinker compresses ViT's $16\times16 + 1~\text{(cls-token)}$ output grid feature tokens into only 5 tokens, thereby saving computational costs.
Specifically, it compress the 2D output of ViT's $16\times16$ into $2\times2$ tokens via  2D average pooling and concatenates it with the cls-token to form $\mathmat{V}\in\mathbb{R}^{5\times D}$.
Consequently, it uses one MLP layer to map the $\mathmat{V}$ to $\mathmat{V'}\in\mathbb{R}^{5 \times D'}$ for modal alignment and  dimension matching of LLM, where $D'$ is the word embedding dimension defined by LLM.
Visual embedding can be inserted into anywhere of input sequence.
Inspired by Shikra~\cite{chen2023shikra}, we employ the digital tokens to represent coordinates, devoid of any vocab, specialty encoders, or any pre-/post-detectors for encoding position information. 
Model variants include Graphist-Tiny utilizing Qwen1.5-0.5B and Graphist-Base engaging 
Qwen1.5-7B.

\subsection{Training Strategy} \label{training}
The proposed Graphist  is trained in three stages. Initially, Stage-1 targets the patch embedding layer and the projector layer. The focal training tasks during this phase predominantly encompass image captioning and the  HLG task, leveraging a substantial batch size coupled with a reduced sequence length for efficiency.
The primary objectives is to calibrate the visual encoder to interpret the alpha channel in RGB-A imagery and to align the projector with both visual and linguistic features.

Progressing to Stage-2, the training expands to encompass both the projector layer and the full LLM. Training tasks from the initial stage are retained; however, the HLG task is administered with greater frequency to highlight the model's fundamental understanding of graphic layout.

Finally, in Stage-3, maintains the focus on the projector layer and the complete LLM, but aiming to adapt the model to a broader range of graphic design task types. Traditional GLG task is also regarded as a specific task type in this stage.
We list primary configurations during training in \tabref{tab:stage}.
\begin{table}[t!]
    \centering
    \renewcommand{\arraystretch}{1.3}
    \renewcommand{\tabcolsep}{1mm}
    \caption{
        \textbf{Implementation details} including training stages, datasets, and essential parameters: ``BS'' for batch size, ``Length'' for total sequence length. "Graphist" is trained on academic data; ``Graphist*'' on proprietary data. ``PE.'' denotes the Patch Embedding Layer.  ``Cap.'' signifies image captioning task. ``Vari.'' denotes the task variants in \secref{sec:cglv2}.
        ``IH-RGBA'' is in-house image-text dataset, where images with alpha channel. ``IH-Design'' is 80k in-house Graphic Design Dataset like Crello~\cite{yamaguchi2021crello}.
    }
    \begin{tabular}{c|llm{1.5cm}m{1cm}m{2.5cm}m{2cm}}
    \hline\toprule
\multirow{2}{*}{Stage} & \multirow{2}{*}{BS} & \multirow{2}{*}{Length} & \multirow{2}{*}{Tune Part} & \multirow{2}{*}{Task} & \multicolumn{2}{c}{Training datasets}  \\
        &   &  &  &     & Graphist  &  Graphist* \\ \hline
      1  &  128 &  1536 & PE \newline Projector  & Cap.\newline HLG        & ShareGPT4v\cite{chen2023sharegpt4v}\newline Flickr30k\cite{plummer2015flickr30k}\newline Crello\cite{yamaguchi2021crello}   &  + IH-RGBA  \\ \hline
      2  &  64  &  2048 & Projector \newline LLM & Cap.\newline HLG        &  ShareGPT4v\cite{chen2023sharegpt4v}\newline Flickr30k\cite{plummer2015flickr30k}\newline Crello\cite{yamaguchi2021crello}  &  + IH-RGBA \newline + IH-Design   \\ \hline
      3  &  64  &  3584 & Projector \newline LLM & Cap.\newline HLG\newline GLG \newline Vari.  &  ShareGPT4v\cite{chen2023sharegpt4v}\newline Flickr30k\cite{plummer2015flickr30k}\newline Crello\cite{yamaguchi2021crello}\newline CGL-V2\cite{li2023relation}  &  + IH-RGBA \newline + IH-Design   \\ \bottomrule
    \end{tabular}
    \label{tab:stage}
    \vspace{-10pt}
\end{table}

To train the model to arrange inputs layer order and spatial coordinates, we randomly shuffle the input elements with a 0.75 probability for all of the Graphic Design datasets used in the training process. 
In these three stages, the first stage trains 10k steps, while the second and third stages both train 20k steps. The training for the Graphist-Base is over the course of 5 days, whereas for the Graphist-Tiny model, a shorter duration of 2 days is anticipated.

\section{Experiment}
\label{sec:experiment}
\subsection{Datasets}
\subsubsection{Crello.~\cite{yamaguchi2021crello}}
Crello dataset \footnote{\url{https://huggingface.co/datasets/cyberagent/crello}} furnishes an array of graphic compositions derived from a web-based design utility, namely Crello (Now change the name to VistaCreate\footnote{\url{https://create.vista.com/}}). 
It covers an extensive range of graphic compositions suited for various applications such as social media infographics, digital banner ads, blog headers, and printed poster templates.
Inside the dataset, each graphic composition includes detailed information about the layering order, spatial positioning, and categorical details of the design elements. The dataset is advantageous for tasks like GLG and has been the foundation for several methods~\cite{yamaguchi2021crello,inoue2023FlexDM}. Additionally, it is a good playground for HLG task.
In Flex-DM~\cite{inoue2023FlexDM}, the dataset is partitioned into 19,095 training, 1,951 validation, and 2,375 testing examples.
However, they used the Crello v2, but since the current version released by Crello is v4, we used the intersection of all parts in the two version test sets, a total of 242 graphic
compositions as the test set in experiments.

\label{sec:cglv2}
CGL-Dataset V2 is an extension of the CGL-Dataset~\cite{DBLP:conf/ijcai/ZhouXMGJX22} and provides a dataset for generating advertising poster layouts, including 60,548 training samples and 1,035 testing samples.
Previous work~\cite{li2023relation} has built two tasks based on this dataset: the Content-Aware task and the Content-Agnostic task. In the Content-Aware task, the model is challenged with an input image and textual content, and is required to output the positions of all textual content in the image while autonomously adding underlay and decorative information. The Content-Agnostic task, on the other hand, only requires an image as input. In this case, the model autonomously determines the output, which includes text, underlay, decoration, and other relevant layering information.
Graphist has integrated these tasks alongside HLG/GLG tasks to facilitate comparison across a broader spectrum of layout methods~\cite{li2019layoutgan,zhou2022composition,kong2022blt,inoue2023layout,li2023relation}.

\subsection{Evaluation Metrics} \label{sec:metrics}
\subsubsection{Inverse order pair ratio (IOPR).}
The appropriate order of layers is crucial for the results of HLG.
We develop IOPR, which is a ratio representing the fraction of overlap element pairs that are in inverse order according to the model's predictions out of all possible overlapping element pairs. 
It is calculated as 
\begin{equation}
\text{IOPR} = \frac{\sum_{i=1}^{n-1} \sum_{j=i+1}^{n} \mathbbm{1} \left(\mathset{O}_j < \mathset{O}_i \land \mathbbm{1}\left(i, j\right) \right)
}{\sum_{i=0}^{n-1} \sum_{j=i+1}^{n} \mathbbm{1}} ,
\end{equation}
where $n$ is the number of layers in the hierarchical structure.
$\mathbbm{1}$ is an indicator function that returns 1 if the argument condition is true and 0 otherwise.
$\mathcal{O}$ denotes the output order or predicted order of the layers as determined by the model. $\mathcal{O}_i$ and $\mathcal{O}_j$ correspond to the predicted order positions of the $i^{th}$ and $j^{th}$ layers, respectively.
$\mathtt{overlap}(i, j)$ is a predicate function that determines whether the $i^{th}$ and $j^{th}$ layers overlap.
A low IOPR would suggest that the model is quite accurate in predicting the correct order of layers because there are few inverse order pairs. Contrastingly, a high IOPR would suggest the model often predicts the wrong sequence, indicating lower prediction accuracy. In our experiments, we employed $\text{IOPR}_{\text{min}}$ and $\text{IOPR}_{\text{avg}}$ to evaluate the performance of the model, representing respectively the average score and the minimum score in the test dataset. 
While the average IOPR across a dataset can provide significant insights, it should be noted that variances in layer orders for specific graphic compositions, when compared to ground truth, do not inherently imply subpar layout performance.

\subsubsection{GPT-4V Eval.}
In addition to the layers order, the overall aesthetic quality and harmony of the elements in the graphic composition are vitally important. We utilize GPT-4V to evaluate our approach and compare it with other alternatives.
Following COLE~\cite{jia2023cole}, we use four scores named GPT-4V rating including $S_{DL}$, $S_{GI}$, $S_{IO}$ and $S_{TV}$ use GPT-4V to evaluate the quality of the graphic composition we generate. 

\begin{itemize}
  \item $S_{DL}$ means the graphic design should present a clean, balanced, and consistent layout. The organization of elements should enhance the message, with clear paths for the eye to follow.
  \item $S_{GI}$ reflects that any graphics or images used should enhance the design rather than distract from it. They should be high quality, relevant, and harmonious with other elements. 
  \item $S_{IO}$ evaluates the innovation level of the design.
  \item $S_{TV}$ represents text readability. A lower score would be assigned if the readability of the text is poor due to the color of the text being similar to the background color or overlapping of the text. 
\end{itemize}

As a supplement to GPT-4V rating~\cite{jia2023cole},
we propose GPT-4V voting. Here,
GPT-4V also partakes in a comparative analysis. It selects the most proficient graphic composition when confronted with two competing outputs. This preference distribution acts as a testament to the discernibility of GPT-4V in recognizing and preferring one method over the other across a range of comparative samples.
GPT-4V rating~\cite{jia2023cole} and GPT-4V voting together form the GPT-4V Eval for evaluating layout ability.

\subsection{Comparison with SoTA}

We trained our model using the approach outlined in \secref{training} and compared it to other state-of-the-art methods. In the following comparison, we represent the model trained only on public datasets as ``Graphist'', while  ``Graphist*'' denotes the model trained on in-house data in addition to public datasets.

\subsubsection{Results in GLG \& HLG tasks.}
\begin{table}[t]
\centering
\renewcommand{\arraystretch}{1.3}
\renewcommand{\tabcolsep}{0.6mm}
 \caption{\textbf{GPT-4V Eval on Crello}. 
 The table demonstrates the performance of different methods on the Crello dataset.
 Graphist is Graphist-Base built upon Qwen1.5-7B. The scores on the left are GPT-4V rating and $\text{IOPR}_\text{avg}$, while the chart on the right showcases a comparative evaluation using GPT-4V voting against the FlexDM~\cite{inoue2023FlexDM}, Gemini-1.5-Pro (Gemini) and GPT-4V. Flex-DM is compared based on the GLG task (a, b), while Gemini-1.5-Pro and GPT-4V are based on the HLG task (c, d).}
\begin{tabular}{
  >{\arraybackslash}m{1.5cm}  
  | > {\centering\arraybackslash}m{1cm} >{\centering\arraybackslash}m{0.9cm} 
  >{\centering\arraybackslash}m{0.9cm}  
  >{\centering\arraybackslash}m{0.9cm} 
  >{\centering\arraybackslash}m{0.9cm}
  >{\centering\arraybackslash}m{0.9cm}| > {\centering\arraybackslash}m{4.1cm} 
}
\toprule
Method  & Task & $S_{DL}$   & $S_{GI}$   & $S_{IO}$   & $S_{TV}$ & $\text{IOPR}$ 
& \multirow{5}{*}{$\includegraphics[scale=0.37]{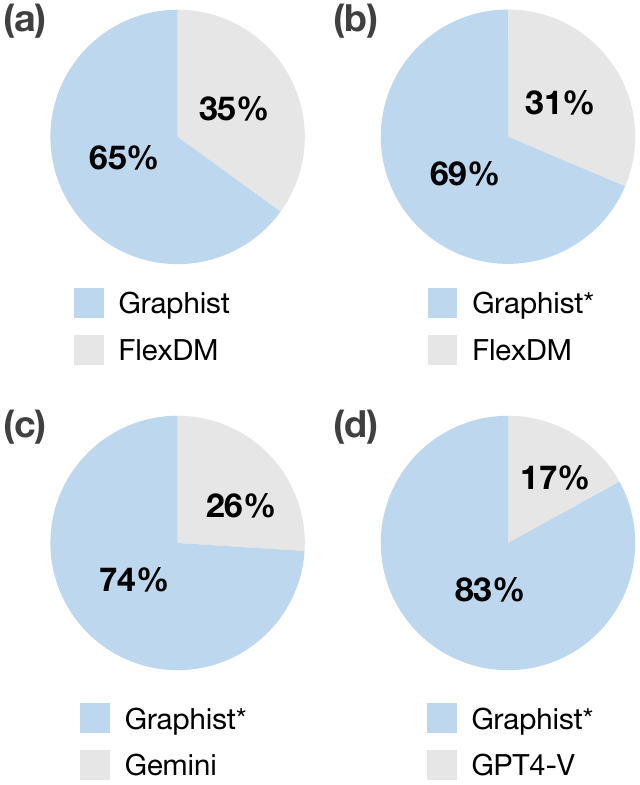}$} \\ \cline{1-7}
Flex-DM & \textit{GLG} & 5.43 & 6.13 & 4.69 & 4.60 & -\\
GPT-4V & \textit{GLG} & 5.10 & 5.83 & 4.54 & 3.84 & -\\
Gemini & \textit{GLG} & 4.87 & 5.62 & 4.39 & 4.09 & -\\
Graphist    & \textit{GLG} & 5.60 & 6.64 & 4.84 & 4.86 & -\\ 
Graphist$*$ & \textit{GLG} & \textbf{5.72} & \textbf{6.70} & \textbf{5.03} & \textbf{5.37} & -\\ \cline{1-7}
GPT-4V & \textit{GLG} & 4.19 & 4.66 & 3.71 & 3.25 & 0.45\\
Gemini & \textit{GLG} & 5.06 & 5.94 & 4.33 & 4.21 & 0.70\\
Graphist    & \textit{HLG} & 5.66 & 6.60 & 5.02 & 4.93 & 0.96& \\
Graphist$*$ & \textit{HLG} & \textbf{5.85} & \textbf{6.90} & \textbf{5.10} & \textbf{5.24} & \textbf{0.97}\\ \toprule
\end{tabular}
\label{tab:crello_res}
\end{table}
The Crello dataset evaluation underscores the dominance of Graphist* and Graphist over Flex-DM across both tasks. In the GLG task, Graphist* achieved the highest scores in three out of four domains, with marked gains in $S_{GI}$, attaining a commanding 6.70, and similarly impressive performance in $S_{DL}$ (5.72) and $S_{IO}$ (5.03). In $S_{TV}$, Graphist* rendered a commendable score, highlighting the method's robustness in maintaining text readability amidst complex designs. 
Shifting to the more challenging HLG task, Graphist* not only continued its strong showing but exhibited even more dramatic enhancements, scoring uniformly higher over the counterpart, Flex-DM. Noteworthy is the peak score of 5.85 for $S_{DL}$, underscoring Graphist*'s proficiency in clean and balanced layout generation. This version also excelled notably in graphic-image synergy with a top score of 6.90 in $S_{GI}$ and innovation with a superior score of 5.10 in $S_{IO}$. Moreover, the $S_{TV}$ metric remains above Flex-DM, reflecting the model's diligence in preserving text clarity across design variations.
We also evaluate the performance of top-tier proprietary linear mixed models, specifically Gemini-1.5-Pro (Gemini) and GPT-4V, on both tasks. Although they offer greater versatility, there is a significant performance gap compared to our Graphist on layout-centric tasks, such as GLG and HLG.
 Regarding the evaluation details, there may be situations where GPT-4V and Gemini-1.5-Pro cannot correctly infer the required JSON format. We have prepared format post-processing for their common return result types, and we give these two methods five inference opportunities for each test case. If all are unsuccessful, then skip. In the end, GPT-4V retained 120 results, and Gemini-1.5-Pro retained 240 results. We reported their scores on these subsets.

\begin{figure}[t]
\centering
\begin{overpic}[width=0.99\textwidth]{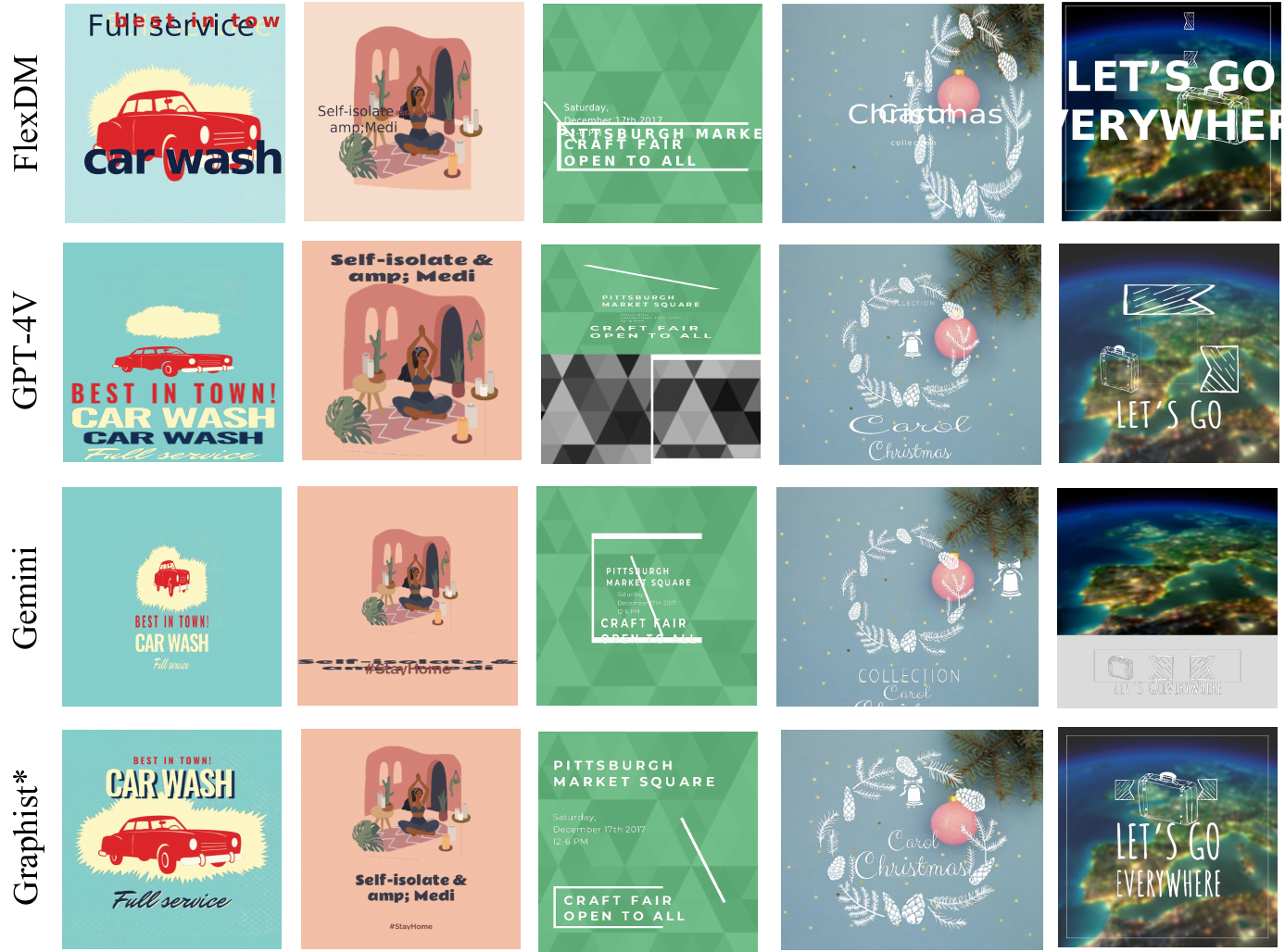}
\end{overpic}
\caption{
\textbf{Results visualization of the GLG task on the Crello dataset.} The results for Flex-DM were derived from their open-source code, whereas the results for GPT-4V and Gemini-1.5-Pro are obtained in zero-shot manner.}
\label{fig:crello_compare}
\end{figure}
\begin{figure}[t]
\centering
\begin{overpic}[width=0.98\textwidth]{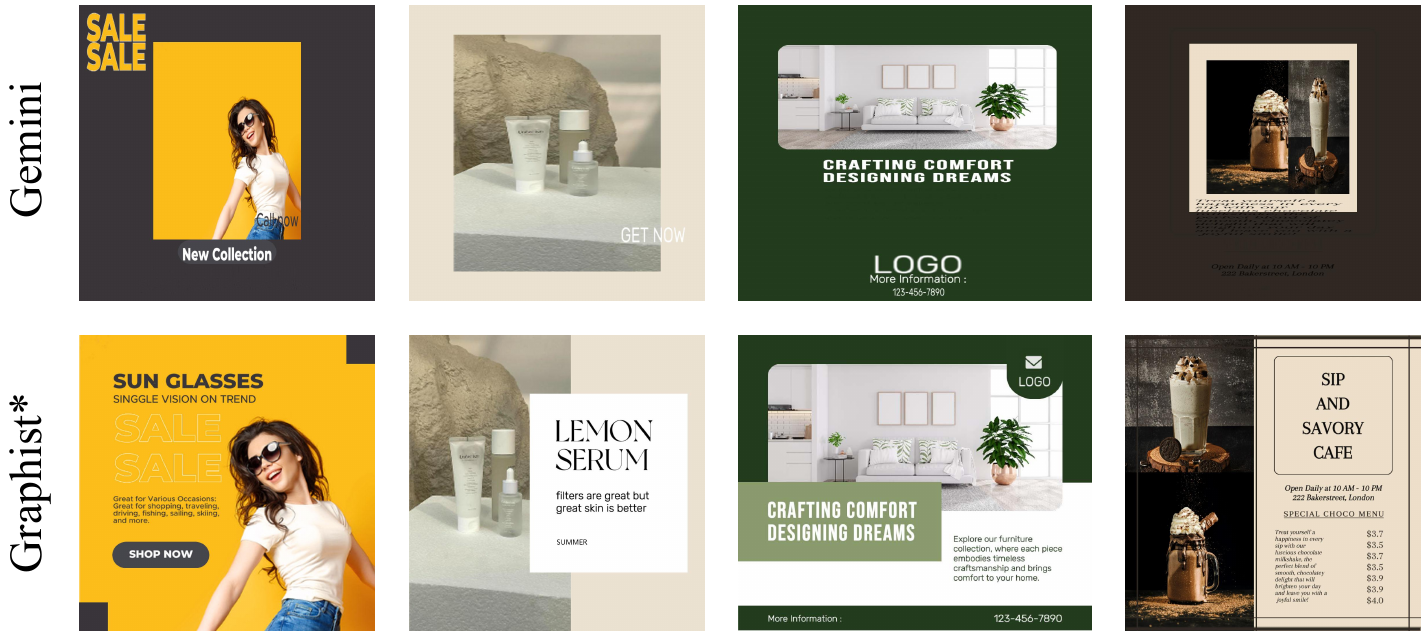}
\end{overpic}
\caption{
\textbf{
Comparison with SoTA method
 on the HLG task with real-world design elements.} 
Both Gemini-1.5-Pro and Graphist models are tasked with HLG using identical real-world design elements. The outcomes indicate superior design quality achieved by our Graphist* when compared with Gemini-1.5-Pro.}
\label{fig:cc_gemini}
\end{figure}

In the visualization of the GLG outcomes in~\figref{fig:crello_compare}, Graphist outperforms competitors including Flex-DM, GPT-4V, and Gemini-1.5-Pro. The other methodologies grapple with challenges such as text overlap and image distortion.
Furthermore, we evaluated real-world design elements, contrasting the performance of Gemini-1.5-Pro with Graphist, as depicted in~\figref{fig:cc_gemini}. For complex design tasks involving over ten layers, our approach yield more refined design outcomes. Although Gemini-1.5-Pro managed to accomplish the HLG task, the design acuity is markedly distinct from that of Graphist.

\subsubsection{Results on CGL-V2.}
\begin{table}[ht]
\centering
\caption{\textbf{Comparison with content-aware methods on CGL-V2.} In this table, we compare Graphist based on Qwen1.5-7B\cite{qwen} against other content-aware methods. Content-aware refers to generating a layout based on the input background image and text content. C-R means Composition-relevant.}
\renewcommand{\arraystretch}{1.3}
\renewcommand{\tabcolsep}{1mm}
\begin{tabular}{>{\arraybackslash}m{3cm}|>{\centering\arraybackslash}m{1.3cm}>{\centering\arraybackslash}m{1.3cm}|>{\centering\arraybackslash}m{1.2cm}>{\centering\arraybackslash}m{1.2cm}>{\centering\arraybackslash}m{1.2cm}}
\toprule
\multirow{2}{*}{Method} & \multicolumn{2}{c|}{C-R measures} & \multicolumn{3}{c}{Graphic measures} \\
    & $R_{com} \downarrow$    & $R_{occ} \uparrow$    &$R_{ali} \downarrow$    & $R_{ove} \downarrow$   & $R_{und} \uparrow$    \\ \midrule
ContentGAN~\cite{li2019layoutgan} & 31.930 & \textbf{1.000} & 0.009 & 0.065 & 0.840 \\
CGL-GAN~\cite{zhou2022composition}   & 16.040 & 0.875 & 0.007 & 0.081 & 0.732 \\
RADM~\cite{li2023relation}      & \textbf{10.260} & 0.997 & 0.008 & 0.046 & 0.983 \\
Graphist*      & 12.150 & \textbf{1.000} & \textbf{0.004} & \textbf{0.002} & \textbf{0.996} \\ \toprule
\end{tabular}
\label{tab:cgl_aware}
\end{table}

\begin{table}[ht]
\centering
\caption{\textbf{Comparison with content-agnostic methods on CGL-V2.} In this table, we compare Graphist based on Qwen1.5-7B\cite{qwen} against other content-agnostic methods. Content-agnostic refers to generating a layout based solely on the background image. C-R means Composition-relevant.}
\renewcommand{\arraystretch}{1.3}
\renewcommand{\tabcolsep}{1mm}
\begin{tabular}{>{\arraybackslash}m{3cm}|>{\centering\arraybackslash}m{1.3cm}>{\centering\arraybackslash}m{1.3cm}|>{\centering\arraybackslash}m{1.2cm}>{\centering\arraybackslash}m{1.2cm}>{\centering\arraybackslash}m{1.2cm}}
\toprule
\multirow{2}{*}{Method} & \multicolumn{2}{c|}{C-R measures} & \multicolumn{3}{c}{Graphic measures} \\
           & $R_{com} \downarrow$    & $R_{occ} \uparrow$    &$R_{ali} \downarrow$    & $R_{ove} \downarrow$   & $R_{und} \uparrow$        \\ \midrule
BLT~\cite{kong2022blt}        & 28.540 & \textbf{1.000} & 0.004 & 0.002 & 0.993 \\
LayoutDM~\cite{inoue2023layout}   & 21.300 & \textbf{1.000} & 0.006 & 0.039 & 0.896 \\
RADM~\cite{li2023relation}       & 10.260 & 0.997 & 0.008 & 0.046 & 0.983 \\
Graphist*  & \textbf{7.074}  & \textbf{1.000} & \textbf{0.0003} & \textbf{0.001} & \textbf{1.000} \\ 
\toprule
\end{tabular}
\label{tab:cgl_agnostic}
\end{table}

\begin{figure}[t]
\centering
\begin{overpic}[width=0.99\textwidth]{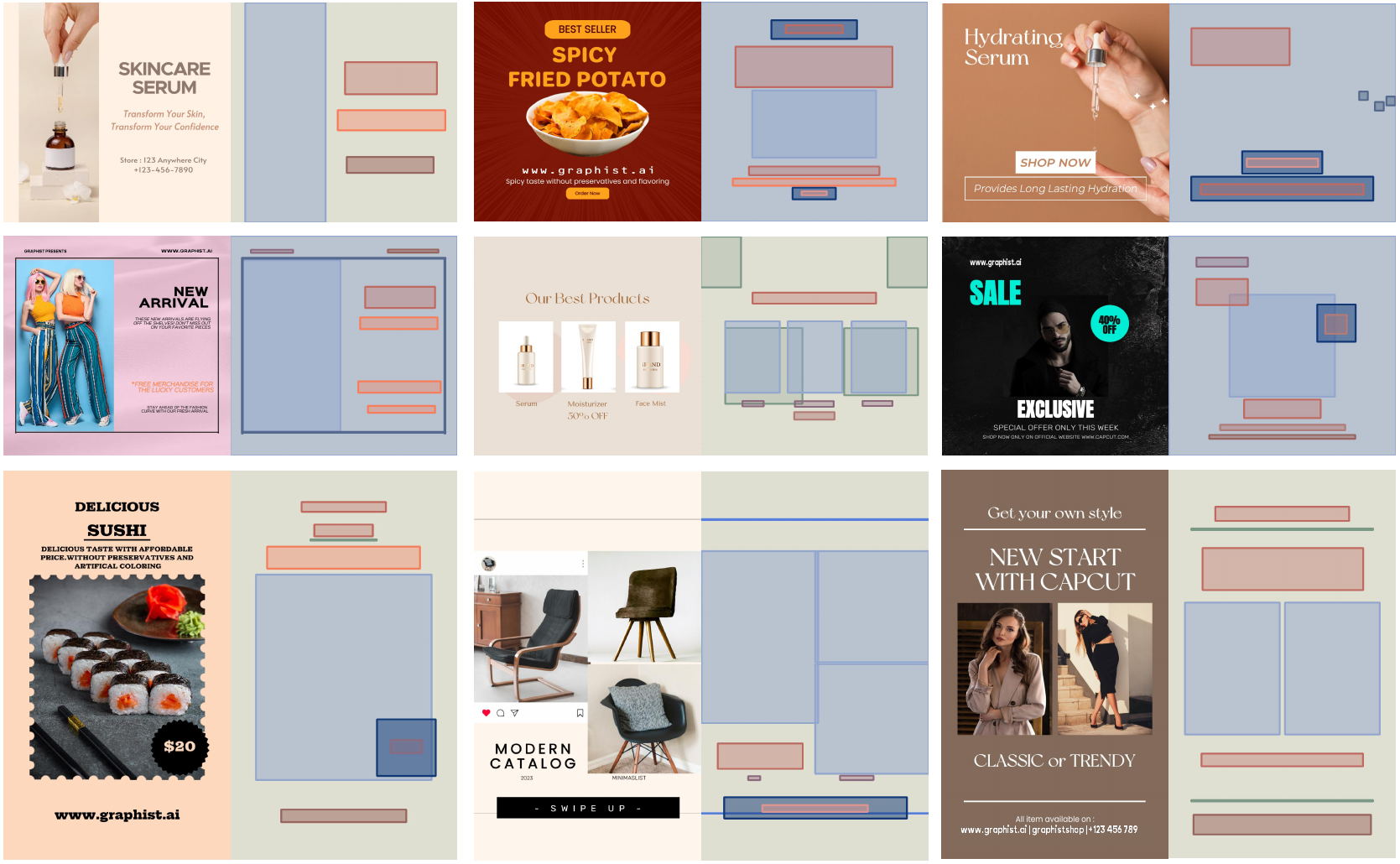}
\end{overpic}
\caption{
\textbf{More user-generated designs via graphist web demo}.
To evaluate usability, we invited numerous non-expert volunteers to upload their design components to our Graphist web demo, resulting in the creation of their own design projects. The image displays a selection of these high-quality outcomes alongside their respective layouts. Within the layout map, various colors signify distinct layer attributes as recognized by the model.
}
\label{fig:user}
\end{figure}
Besides the HLG and GLG tasks, our model can adapt to various other design tasks through flexible configurations. We conducted Content-Aware and Content-Agnostic layout generation experiments on the CGL-V2 dataset based on Qwen1.5-7B and following the experiment settings and evaluation metrics of~\cite{zhou2022composition}.

In the Content-Aware layout design, from~\tabref{tab:cgl_aware}, Graphist stands out with the lowest $R_{com}$ of 12.150, the highest $R_{occ}$ tied with ContentGAN at 1.000, and leading performances in all graphic measures ($R_{ali} = 0.004, R_{ove} = 0.002, R_{und} = 0.996$), indicating superior grapic visual balance, non-emptiness, alignment, overlap avoidance, and underlay success.

Similarly, within Content-Agnostic layout task, Graphist again showcases its robust capability by achieving the best $R_{com}$ score of 7.074 and perfect $R_{occ}$ score, which is illustrated in~\tabref{tab:cgl_agnostic}. It also surpasses competitors in graphic measures, underpinning its strength in alignment ($R_{ali} = 0.0003$), minimizing elements overlap ($R_{ove} = 0.001$), and maximizing the informative underlay usage ($R_{und} = 1.0$). These results cement Graphist as a versatile and potent solution for automatic layout generation, exhibiting notable advancements in synthesizing information-rich graphic compositions.


\subsection{Ablation Studies}
In this section, we delve into a series of ablation studies to examine the underlying factors that influence the layout quality of Graphist. The specific aspects under investigation include the sequencing of input design elements, the choice of language model, the number of visual tokens, and the encoding of transparency through RGB-A image channels.

\subsubsection{Influence of input layer sequencing.}
\label{abl_layerorder}
In~\tabref{tab:crello_res}, we show the performance comparison of our method with structured versus unstructured input sequences. The findings indicate that our approach achieves better results in the metrics $S_{DL}$, $S_{GI}$, and $S_{IO}$ when dealing with an unordered input. The flexibility introduced by allowing randomness in input ordering contributes to the superior performance observed in $S_{DL}$ and $S_{IO}$ metrics versus a structured sequence. Nevertheless, this unordered approach increases the complexity of text layer placement, reflected by a reduced score in $S_{TV}$.

\subsubsection{Impact of LLM.}
\label{abl_model}
\begin{table}[t]
\centering
\renewcommand{\arraystretch}{1.3}
\renewcommand{\tabcolsep}{1mm}
\caption{\textbf{Different LLM model}. Comparison of Qwen1.5-0.5B, Qwen1.5-7B, and InternLM2-7B performance on the Crello dataset.}
\begin{tabular}{
  >{\arraybackslash}m{2.3cm}  
  | > {\centering\arraybackslash}m{1cm}>{\centering\arraybackslash}m{1cm} 
  >{\centering\arraybackslash}m{1cm}  
  >{\centering\arraybackslash}m{1cm} 
  | > {\centering\arraybackslash}m{1.5cm}>{\centering\arraybackslash}m{1.5cm} 
}
\toprule
LLM  & $S_{DL}$   & $S_{GI}$   & $S_{IO}$   & $S_{TV}$ & $\text{IOPR}_{\text{min}}$ & $\text{IOPR}_{\text{avg}}$\\ \midrule
Qwen1.5-0.5B & 5.44 & 6.35 & 4.94 & 4.60 & 0.641 & 0.963 \\
Qwen1.5-7B  & 5.85 & 6.90 & 5.10 & 5.37 & 0.667 & 0.971 \\ 
InternLM2-7B & 5.60 & 6.50 & 4.90 & 5.00 & 0.476 & 0.966 \\ \toprule

\end{tabular}
\label{tab:crello_modelsize}
\end{table}

\begin{table}[ht]
\centering
\caption{\textbf{Different LLM model.} Comparison of Qwen1.5-7B and InternLM2-7B performance on the CGL-V2 dataset.}
\renewcommand{\arraystretch}{1.3}
\renewcommand{\tabcolsep}{1mm}
\begin{tabular}{>{\arraybackslash}m{3cm}|>{\arraybackslash}m{1.3cm}|>{\centering\arraybackslash}m{1.3cm}>{\centering\arraybackslash}m{1.3cm}|>{\centering\arraybackslash}m{1.2cm}>{\centering\arraybackslash}m{1.2cm}>{\centering\arraybackslash}m{1.2cm}}
\toprule
\multirow{2}{*}{LLM} & \multirow{2}{*}{Task} & \multicolumn{2}{c|}{C-R measures} & \multicolumn{3}{c}{Graphic measures} \\
   & & $R_{com} \downarrow$    & $R_{occ} \uparrow$    &$R_{ali} \downarrow$    & $R_{ove} \downarrow$   & $R_{und} \uparrow$    \\ \midrule
InternLM2-7B & Aware & 12.803 & \textbf{1.000} & \textbf{0.004} & \textbf{0.002} & 0.988 \\
Qwen1.5-7B   & Aware& 12.150 & \textbf{1.000} & \textbf{0.004} & \textbf{0.002} & \textbf{0.996} \\ \midrule
InternLM2-7B  &Agnostic  & 9.510  & \textbf{1.000} & 0.004 & \textbf{0.001} & 0.999 \\ 
Qwen1.5-7B & Agnostic&\textbf{7.074}  & \textbf{1.000} & \textbf{0.0003} & \textbf{0.001} & \textbf{1.000} \\ \toprule
\end{tabular}
\label{tab:abl_cgl}
\end{table}
In this evaluation, we explore the influence of varying LLM configurations on our layout generation results. Similarly, we carry out experiments using the Crello dataset and the CGL-V2 dataset. We evaluate models with differing capacities: Qwen1.5-0.5B, Qwen1.5-7B, and InternLM2-7B. As demonstrated in~\tabref{tab:crello_modelsize} and~\tabref{tab:abl_cgl}, our experiments underline that within the domain of our tasks, the larger models consistently outperform their smaller counterparts. In addition, different models of the same size also reflect performance differences.

\begin{table}[t]
\centering
\renewcommand{\arraystretch}{1.3}
\renewcommand{\tabcolsep}{1mm}
\caption{\textbf{Different visual token length}. Comparison of different sequence length performance on the Crello dataset based on Graphist-Tiny. From our observation, five tokens are sufficient for layout generation. }
\begin{tabular}{
  >{\arraybackslash}m{1.5cm}  
  | > {\centering\arraybackslash}m{1cm}>{\centering\arraybackslash}m{1cm} 
  >{\centering\arraybackslash}m{1cm}  
  >{\centering\arraybackslash}m{1cm} 
  | > {\centering\arraybackslash}m{1.5cm}>{\centering\arraybackslash}m{1.5cm} 
}
\toprule
Length  & $S_{DL}$   & $S_{GI}$   & $S_{IO}$   & $S_{TV}$ & $\text{IOPR}_{\text{min}}$ & $\text{IOPR}_{\text{avg}}$ \\ \midrule
5 tokens & 5.44 & 6.35 & 4.94 & 4.60 & 0.641 & 0.963\\
17 tokens & 5.47 & 6.28 & 4.88 & 4.67 & 0.667 & 0.965\\ \toprule
\end{tabular}
\label{tab:crello_length}
\end{table}

\begin{table}[ht]
\centering
\renewcommand{\arraystretch}{1.3}
\renewcommand{\tabcolsep}{1mm}
\caption{\textbf{Different input channel}. Comparison of RGB input and RGB-A input performance on the Crello dataset. The results indicate that RGB-A performs better, particularly in $S_{IO}$ and $S_{TV}$. This experiment based on Graphist-Tiny model.}
\begin{tabular}{
  >{\arraybackslash}m{1.5cm}  
  | > {\centering\arraybackslash}m{1cm}>{\centering\arraybackslash}m{1cm} 
  >{\centering\arraybackslash}m{1cm}  
  >{\centering\arraybackslash}m{1cm} 
  | > {\centering\arraybackslash}m{1.5cm}>{\centering\arraybackslash}m{1.5cm} 
}
\toprule
Method  & $S_{DL}$   & $S_{GI}$   & $S_{IO}$   & $S_{TV}$ & $\text{IOPR}_{\text{min}}$ & $\text{IOPR}_{\text{avg}}$\\ \midrule
RGB-A & 5.44 & 6.35 & 4.94 & 4.60 & 0.641 & 0.963\\
RGB  & 5.24 & 6.14 & 4.62 & 4.22 & 0.502 & 0.951\\ \toprule
\end{tabular}
\label{tab:crello_rgbargb}
\end{table}

\subsubsection{Influence of visual token length.}
\label{abl_seqlength} In our method, we typically represent an image using a sequence of 5 visual tokens. To evaluate the effects of increased token quantity, potentially enhancing the image representation's granularity, we conducted experiments using a 17 tokens format. The composition of 17 tokens includes 1 cls token and 16 visual tokens, achieved by using a $4\times4$ pooling kernel size. The results presented in~\tabref{tab:crello_length} suggests that lengthening the visual token sequence does not necessarily lead to performance improvements. According to the outcomes, the representation of an image with a quintet of tokens sufficiently captures the necessary information for this specific task.

\subsubsection{RGB \vs RGB-A.}
To assess the effect of omitting the alpha channel, we have also tested the model with traditional three-channel RGB images. As illustrated in~\tabref{tab:crello_rgbargb}, inputs with the additional alpha channel (RGB-A) yield higher-quality outputs with more accurate layer ordering when compared to RGB inputs. The alpha channel provides detailed information that aids the model in discerning textural elements and gradients within the image layers. Importantly, when processing text layers, the alpha channel allows the model to isolate text from potentially distracting backgrounds, facilitating improved clarity and precision in text placement.

\subsection{Real-World Evaluation}
To assess the real-world applicability of our method, we engaged a group of non-expert volunteers to submit their design assets via our Graphist web demo and chose a selection of representative outputs for analysis. As illustrated in~\figref{fig:user}, our model successfully generates visually appealing and cohesive designs across varying canvas dimensions and design elements.

\subsection{Limitations \& Negative Impact}
In order to create an efficient design assistant that adheres to user intent, Graphist only represents a starting point. The automatic generation of complete sets of high-quality materials that meet design intent requirements, and the generation of designs that more closely align with human aesthetic preferences, are areas that require further exploration. As we explore more intelligent graphic design systems, potential negative consequences primarily involve generating homogeneous design results and the environmental impact associated with the carbon consumption of model training.


\section{Conclusion}
\label{sec:conclusion}
This paper represents a step forward from traditional graphic layout generation by introducing the hierarchical layout generation task, which enhances graphic design automation by effectively handling disordered design elements, thereby increasing creative potential and efficiency. To address this more challenging task, we proposed Graphist, a novel LMM that tackles HLG tasks as sequence generation challenges. Graphist takes RGB-A images as input and produces JSON draft protocols that define the layout parameters of graphic compositions. To appropriately evaluate HLG tasks, we introduced two metrics: the Inverse Order Pair Ratio and GPT-4V Eval. Our evaluation metrics demonstrate that Graphist achieves state-of-the-art results, providing a strong baseline for generating automated graphic designs that are more creative and diverse.

%
%
\bibliographystyle{splncs04}
\bibliography{egbib}
\end{document}